\documentclass[letterpaper, 10 pt, conference]{ieeeconf}  
\usepackage{amsmath} 
\usepackage{adjustbox} 
\usepackage{graphicx} 
\usepackage{booktabs} 
\usepackage{multirow} 
\usepackage{amsmath} 
\usepackage{amssymb} 
\usepackage{siunitx}

\IEEEoverridecommandlockouts                              

\usepackage{amssymb}
\usepackage{amsmath}
\usepackage{algorithm}
\usepackage{algpseudocode}
\usepackage{graphicx}
\usepackage{bm}
\usepackage{array,booktabs}

\title{\LARGE \bf
Heteroscedastic Bayesian Optimization-Based Dynamic PID Tuning for Accurate and Robust UAV Trajectory Tracking
}

\author{Fuqiang Gu$^{1}$, Jiangshan Ai$^{1}$, Xu Lu$^{1}$, Xianlei Long$^{1}$, Yan Li$^{2}$, Tao Jiang$^{3}$, Chao Chen$^{1}$, Huidong Liu*$^{1}$
\thanks{*This paper is supported by the National Natural Science Foundation of China (No. 42174050, No. 42474027, No. 62403085), Graduate research and innovation foundation of Chongqing, China (No. CYS25044), Chongqing Innovative and Entrepreneurial Program of Returned Overseas Chinese Scholars (No. cx2021047), and Chongqing Startup Program for Doctorate Scholars (No. CSTB2022BSXM-JSX005). \textit{Corresponding author: Huidong Liu}.}
\thanks{$^{1}$H. Liu, J. Ai, X. Lu, X. Long, C. Chen and F. Gu are with the College of Computer Science, Chongqing University, Chongqing, China.
        {\tt\small e-mails: \{liuhuidong, ajs, lux\}@stu.cqu.edu.cn, \{ xianlei.long, cschaochen, gufq\}@cqu.edu.cn}}%
\thanks{$^{2}$Y. Li is with the School of Computing, Macquarie University, Sydney, Australia.
        {\tt\small e-mail: y.li@mq.edu.au}}%
\thanks{$^{3}$T. Jiang is with the School of Automation, Chongqing University, Chongqing, China. 
 {\tt\small e-mail: jiangtao\_1992@cqu.edu.cn}}
}
\begin{document}

\maketitle
\thispagestyle{empty}
\pagestyle{empty}

\begin{abstract}
Unmanned Aerial Vehicles (UAVs) play an important role in various applications, where precise trajectory tracking is crucial. However, conventional control algorithms for trajectory tracking often exhibit limited performance due to the underactuated, nonlinear, and highly coupled dynamics of quadrotor systems. To address these challenges, we propose HBO-PID, a novel control algorithm that integrates the Heteroscedastic Bayesian Optimization (HBO) framework with the classical PID controller to achieve accurate and robust trajectory tracking. By explicitly modeling input-dependent noise variance, the proposed method can better adapt to dynamic and complex environments, and therefore improve the accuracy and robustness of trajectory tracking. To accelerate the convergence of optimization, we adopt a two-stage optimization strategy that allow us to more efficiently find the optimal controller parameters. Through experiments in both simulation and real-world scenarios, we demonstrate that the proposed method significantly outperforms state-of-the-art (SOTA) methods. Compared to SOTA methods, it improves the position accuracy by 24.7\% to 42.9\%, and the angular accuracy by 40.9\% to 78.4\%.

\end{abstract}

\section{Introduction}
Recently, UAVs have demonstrated significant potential in various fields, such as logistics~\cite{du2024ai}, environmental monitoring~\cite{motlagh2023unmanned}, agricultural management~\cite{lee2023single}, disaster relief~\cite{gui2024coverage}, and infrastructure inspection~\cite{petit2024moar}, due to their flexibility and efficiency. Among these applications, precise and robust trajectory tracking is crucial for UAVs to successfully accomplish their tasks~\cite{zou2024model, liu_APF_uav}. However, as a typical underactuated system, the quadrotor UAV is characterized by highly nonlinear and strongly coupled dynamics, making it challenging to achieve accurate and arbitrary trajectory tracking.

To address the above challenge, one needs to develop precise and robust UAV control algorithms. While the classical Proportional-Integral-Derivative (PID) control~\cite{lopez2023pid} is widely used for stabilizing UAV attitude and trajectory tracking due to its simplicity and efficiency, it often struggles with nonlinear dynamics and external disturbances. Besides, it requires manual tuning for optimal performance, which is challenging in dynamic environments.

To improve the stability, precision, and robustness of UAV controllers under complex and dynamic environments, some researchers have turned to model-based controllers. For instance, Lindqvist et al.~\cite{lindqvist2020nonlinear} proposed a novel Nonlinear Model Predictive Control (NMPC) method for UAV navigation and dynamic obstacle avoidance.  Minařík et al. ~\cite{minavrik2024model} introduced a real-time control architecture based on model predictive path integral (MPPI) control for agile UAVs, aiming to solve the problem of efficient flight in obstacle-rich environments. However, these methods are often computationally expensive, and rely heavily on accurate system models, limiting applicability in uncertain environments.

In recent years, some learning-based control methods have been developed for trajectory tracking~\cite{yu2024modular, huang2023datt}, which can adapt to complex, unstructured environments and reduce dependency on explicit system model. For example, Han et al.~\cite{han2022cascade} proposed a cascade flight control method for quadrotors based on deep reinforcement learning, which leveraged a layered control architecture to enhance stability, robustness, and trajectory tracking performance in dynamic environments. Through large-scale training, these methods demonstrate strong adaptability and robustness, making them increasingly effective in addressing complex control challenges. However, learning-based methods typically require substantial computational resources and are heavily dependent on a large amount of training data. This makes it difficult for them to quickly adapt to different trajectory requirements or environmental changes. 

To improve the efficiency and reduce the dependence on the huge training data, recent works have integrated traditional PID controllers with nonlinear dynamic compensation mechanisms. For instance, Noordin et al.~\cite{noordin2021adaptive} proposed an adaptive PID controller based on Sliding Mode Control (SMC) to address nonlinearity and parameter uncertainties in quadrotor UAV stabilization. Building on this, Berkenkamp et al.~\cite{berkenkamp2016safe} introduced a PID parameter optimization method using BO with Gaussian Process (GP) modeling, significantly enhancing the control performance. To further improve adaptability, Zhao et al.~\cite{zhao2024domain} developed an enhanced BO algorithm that optimizes modeling efficiency, demonstrating greater robustness in dynamic environments. However, these methods depend heavily on modeling data heterogeneity, and their performance may degrade in scenarios with complex data distributions or high noise levels, posing a critical challenge for future research.

In this study, we propose HBO-PID, a novel control method for trajectory tracking that integrates the classical PID controller with HBO. Compared to existing methods, HBO-PID features higher efficiency, lower pose error, better robustness, and adaptability to dynamic and complex environments. This is mainly attributed to the introduction of HBO, which extends traditional BO by explicitly modeling input-dependent noise variance (heteroscedasticity). Instead of assuming a constant noise level throughout the search space, HBO uses a surrogate model, often a GP, to estimate both the mean and the noise variance at each input point. This allows the algorithm to adapt to regions with varying uncertainty, capturing scenarios in which some areas of the search space are noisier than others. To accelerate the convergence and robustness of optimization, we utilize a two-stage optimization strategy. In the first phase, the algorithm focuses on large-scale error adjustment to help the system find a reasonable controller parameter range; while in the second phase, it further refines the optimization to reduce the remaining errors and enhance the robustness of the controller. In this way, HBO-PID can accurately and efficiently perform trajectory tracking even in dynamic and complex environments. 

In summary, the main contributions of this study are as follows:
 \begin{itemize}
    \item We propose HBO-PID, a novel control method for accurate and robust trajectory tracking. Different existing methods, which either sacrifice computational cost or require a huge amount of training data, HBO-PID can accurately and robustly conduct trajectory tracking without pre-training. 
     \item We propose to use the HBO to better model the tracking noise regardless of inputs from dynamic and complex environments. This allows the UAV to adapt to noisy input and changing environments. 
     \item We adopt a two-stage optimization strategy, which avoids the computational cost of fine-tuning too early and ensures the optimal selection of controller parameters. This helps improve the accuracy and robustness of trajectory tracking. 
     \item We evaluate the proposed method through both simulation and real-world experiments. Experimental results show that HBO-PID significantly outperforms state-of-the-art methods in terms of pose error and robustness. 
\end{itemize}


\section{Problem Statement and Preliminary}
\subsection{Problem Statement}
Let us consider a UAV operating in a 3-dimensional bounded environment and moving along a trajectory. Its dynamic behavior is described using a discrete-time state-space model, which is expressed as the following equation:
\begin{equation}
    \mathbf{x}(t+1)=\mathbf{f}(\mathbf{x}(t),\mathbf{u}(t))+\mathbf{w}(t),
\end{equation}
where $\mathbf{x}(t)$, $\mathbf{u}(t)$, and $\mathbf{w}(t)$ are the state vector, control input vector, and heteroscedastic process noise of the quadrotor at time $t$, respectively, with the variance of $\mathbf{w}(t)$ potentially varying over time or state.

The objective of trajectory tracking is to design a control law $\mathbf{u}(t)$ that stabilizes the quadrotor while minimizing the following cost function:
\begin{equation}
    J=\sum_{t=0}^T\left(\|\mathbf{x}(t)-\mathbf{x}^*(t)\|_Q^2+\|\mathbf{u}(t)\|_R^2\right),
\end{equation}
where $\mathrm{x}^*(t)$ is the desired state trajectory, $Q\in\mathbb{R}^{n\times n}$ and $R\in\mathbb{R}^{m\times m}$ are positive definite weighting matrices that penalize deviations from the desired state and excessive control effort, respectively. In this context, $n$ and $m$ denote the dimensions of the state vector and the control input vector. 

Consider a system consisting of \(\bar{k}\) cascaded PID controllers, where the parameter set for each controller is defined as \(\boldsymbol{\xi}_i = \{K_{p,i}, K_{i,i}, K_{d,i}\}\). The complete parameter set for the entire system can then be expressed as:
\begin{equation}
 \boldsymbol{\Xi} = \{\boldsymbol{\xi}_1, \boldsymbol{\xi}_2, \dots, \boldsymbol{\xi}_{\bar{k}}\}.
\end{equation}

The goal of this study is to determine the optimal parameter set \(\boldsymbol{\Xi}^*\) that minimizes a predefined objective function \(f(\boldsymbol{\Xi})\), formulated as:
\begin{equation}
\boldsymbol{\Xi}^* = \arg \min_{\boldsymbol{\Xi}} f(\boldsymbol{\Xi}).
\end{equation}

\subsection{Quadrotor Dynamic System}
We model the quadrotor as a rigid body controlled by four motors, with its state vector defined as $\mathbf{x} = [\mathbf{p}, \mathbf{q}, \mathbf{v}, \boldsymbol{\omega}]^T$, where $\mathbf{p} \in \mathbb{R}^3$ represents the position of the quadrotor in the inertial frame, $\mathbf{v} \in \mathbb{R}^3$ denotes its velocity in the inertial frame, $\mathbf{q} \in \mathbb{H}$ describes the attitude of the quadrotor using a unit quaternion with $\|\mathbf{q}\|=1$, $\boldsymbol{\omega} \in \mathbb{R}^3$ indicates the angular velocity in the body frame. We use the quadrotor dynamics proposed in Newton-Euler formalism, which describes the translational and rotational motion of the system under the influence of external forces and torques~\cite{han2022cascade}. Its dynamic system is summarized as:
\begin{equation}
\begin{split}
\dot{\mathbf{p}} &= \mathbf{v}, \\
m\dot{\mathbf{v}} &= \mathbf{R}\mathbf{F} + m\mathbf{g}, \\
\dot{\boldsymbol{\mathbf{q}}} &= \frac{1}{2}\boldsymbol{\mathbf{q}}\otimes[\boldsymbol{0},\boldsymbol{\omega}]^T, \\
 \boldsymbol{M} &= \mathbf{J}\dot{\boldsymbol{\omega}} +\boldsymbol{\omega} \times \mathbf{J}\boldsymbol{\omega},
\end{split}
\end{equation}
where $m$ is the mass, $\mathbf{g} = [0, 0, -g]^T$ denotes the gravitational acceleration vector, The symbol $\otimes$ denotes quaternion multiplication (Hamilton product). $\mathbf{R}$ is the rotation matrix derived from the unit quaternion $\mathbf{q}$, and $\mathbf{J}$ represents the inertia tensor. The total thrust $\mathbf{F}_t$ is given by:
\begin{equation}
\mathbf{F}_t = C_f \sum_{i=1}^{4} \boldsymbol{\Omega}_i^2 \mathbf{e}_3,
\end{equation}
where $C_f$ is the thrust coefficient, $\boldsymbol{\Omega_i^2}$ represents the rotational speed of the i-th rotors, $\mathbf{e}_3 = [0, 0, 1]^T$ represents the $z$-axis direction of the body frame.

The total moment $\boldsymbol{M}$ is given by:
\begin{equation}
\boldsymbol{M} = \sum_{i=1}^{4} \left( C_\tau \boldsymbol{\Omega}_i^2 \mathbf{e}_3 + \mathbf{r}_i \times (C_f \boldsymbol{\Omega}_i^2 \mathbf{e}_3) \right),
\end{equation}
where $C_\tau$ is the moment coefficient, $\mathbf{r}_i$ is the position vector of the $i$-th rotor relative to the UAV's center of mass.

\begin{figure*}[thpb]
    \centering
    \includegraphics[width=0.9\linewidth]{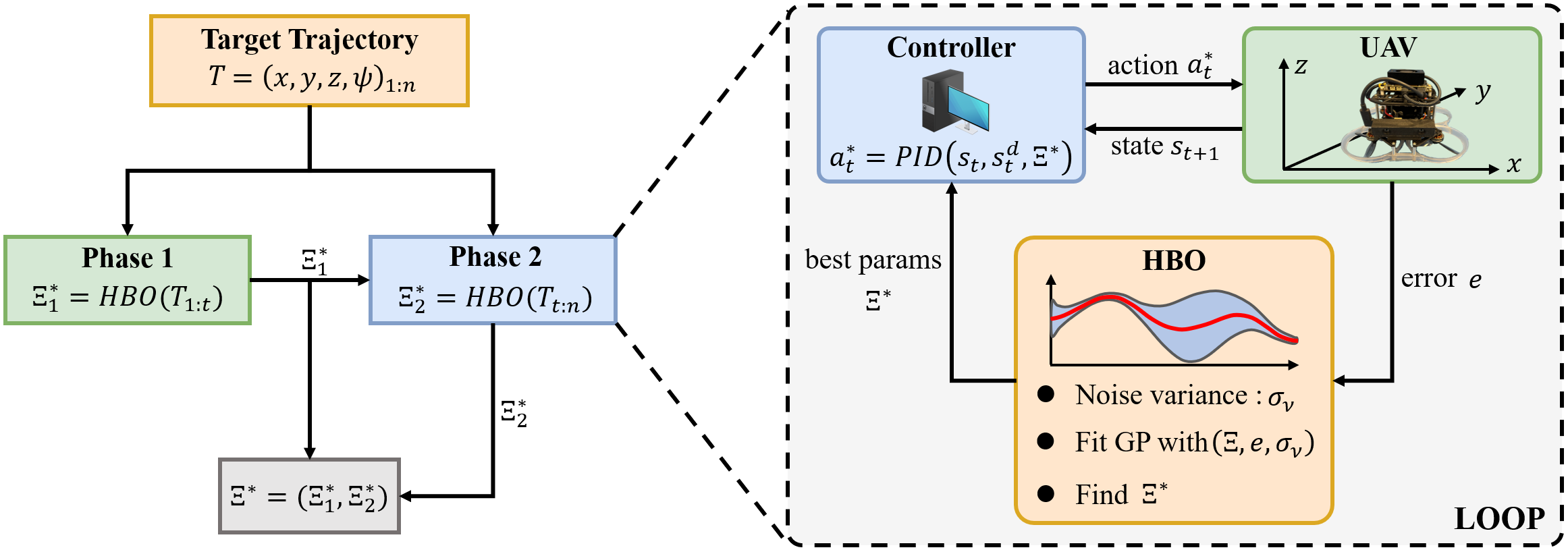}
    \caption{Overall algorithm framework and controller parameter optimization, we first divide the trajectory tracking into two phases and use heteroskedastic Bayesian optimization to search for the optimal PID parameters in each phase to adaptively obtain the best parameter configuration for the current trajectory.}
    \label{fig:algorithm}
\end{figure*}

\subsection{PID Cascade Control}
PID Cascade Control~\cite{lopez2023pid} is an advanced control strategy that employs a hierarchical structure of multiple PID controllers to achieve high-precision regulation of complex systems. Its core architecture is composed of two distinct levels: the outer loop controller, which handles position control, and the inner loop controller, which manages attitude control~\cite{lee2010geometric}. The outer loop controller is responsible for generating the desired thrust $f^d$ and desired angles $\mathbf{\Theta}^d = [\varphi^d, \theta^d, \psi^d]$.


The inner loop controller generates the desired control torque $\tau^d$ based on the output of the outer loop controller. Each controller adopts the following form of PID control:
\begin{equation}
\label{formu:pid}
    u(t) = K_p e(t) + K_i \int e(t) dt + K_d \frac{de(t)}{dt},
\end{equation}
where $e(t)$ represents the error, and $K_p, K_i, K_d$ are the control gain parameters.

This architecture provides a reliable foundational control scheme for the subsequent integration of heteroscedastic optimization algorithms.


\subsection{Bayesian Optimization}
In trajectory tracking control of UAVs, optimizing control parameters presents significant challenges due to high-dimensional search spaces, nonlinear objective functions, and computationally expensive evaluations. The objective function, defined as an implicit mapping between tracking error and controller parameters ($e = f(\Xi)$), lacks an analytical form and requires full dynamic simulations for evaluation. Traditional gradient-based methods, which rely on explicit derivatives, are impractical under these conditions. BO emerges as a powerful framework for such black-box optimization problems, efficiently balancing global exploration and local exploitation through probabilistic surrogate modeling and strategic sampling.  

Central to BO is the GP, a nonparametric Bayesian regression model that defines a distribution over functions. A GP is fully characterized by its mean function $m(x)$ and covariance kernel $k(x, x')$, which jointly capture prior assumptions about the function’s smoothness and quantify prediction uncertainty at unobserved points:  
\begin{equation}
\label{formu:gp}
    g(x) \sim \mathcal{GP}\left(m(x),\, k(x, x')\right).
\end{equation}

Given a set of noisy observations $\{(x_i, y_i)\}_{i=1}^n$ with $y_i = f(x_i) + \epsilon_i$ and $\epsilon_i \sim \mathcal{N}(0, \sigma_{\nu}^2)$, the GP provides a posterior distribution over the function space. For new test points $X_*$, the joint distribution of observed outputs $\mathbf{y}$ and predicted values $\mathbf{f}_*$ follows:  
\begin{equation}
\label{formu:gp_predict}
    \begin{bmatrix}
\mathbf{y} \\
\mathbf{f}_*
\end{bmatrix}
\sim \mathcal{N}
\left(
\begin{bmatrix}
\mathbf{0} \\
\mathbf{0}
\end{bmatrix},\,
\begin{bmatrix}
K(X,X)+\sigma_\nu^2 I & K(X,X_*)\\
K(X_*,X) & K(X_*,X_*)+\sigma_\nu^2 I
\end{bmatrix}
\right),
\end{equation}  
where $K(X, X)$ and $K(X_*, X_*)$ denote covariance matrices of training and test points, and $\sigma_\nu^2$ denotes the observation noise variance. Traditionally, BO treats this variance as constant (homoscedasticity). However, in UAV control scenarios, where uncertainty can fluctuate with system states or environmental disturbances, this assumption may be overly simplistic. To address this, we propose a method to model noise variance dynamically and integrate it into BO.

\section{Proposed Method}
The proposed method for trajectory tracking is developed upon the classical PID control enhanced by HBO. It consists of two key parts: heteroscedastic noise model and optimization strategy. The overall algorithm framework and the detailed controller parameter optimization process are shown in Fig.~\ref{fig:algorithm}. The left figure shows the overall framework, where the target trajectory is used in stage 1 and stage 2 to optimize the controller parameter $\boldsymbol{\Xi}^*$. In stage 1, the parameter $\boldsymbol{\Xi}^*_1$ is optimized, and in stage 2, $\boldsymbol{\Xi}^*_2$ is further refined. The final optimal parameter $\boldsymbol{\Xi}^*$ is obtained by combining the two stages. The right figure illustrates the specific controller parameter optimization process in each stage, using HBO to find the optimal parameter $\boldsymbol{\Xi}^*$.

\subsection{Heteroscedastic Bayesian Optimization}
\begin{figure}[t]
    \centering
    \includegraphics[width=1.0\linewidth]{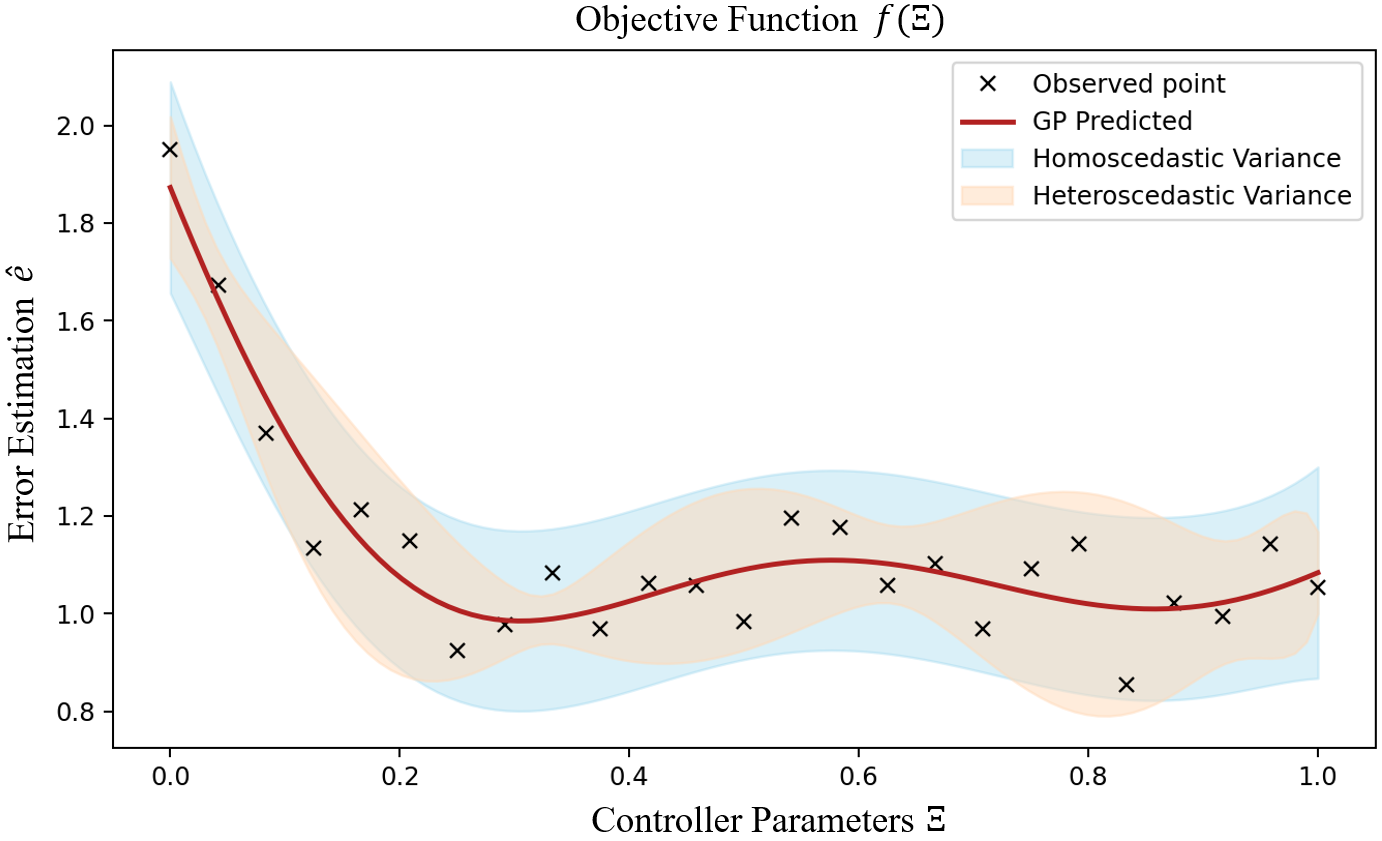}
    \caption{Example of error estimation $\hat{e}$ with Gaussian Process regression, showing predicted error with homoscedastic and heteroscedastic variances.}
    \label{fig:error}
\end{figure}

The homoscedastic noise assumption in traditional BO is often unrealistic in many real-world scenarios. As illustrated in Fig.~\ref{fig:error}, the variability in environmental noise, which originates from nonlinear system dynamics, external disturbances, and inconsistent sensor measurements ~\cite{berkenkamp2016safe, zhao2024domain}, gives rise to pronounced heteroscedasticity. Therefore, traditional BO may not adequately capture the true uncertainty in UAV tracking performance, thereby motivating the adoption of heteroscedastic Bayesian optimization.

To address this challenge, we introduce a heteroscedastic noise model that adapts to varying noise levels across different regions of the search space. We begin by assuming that the noise at distinct input points is independent, which implies that there is no correlation between the noise at different points. From eq.\ref{formu:gp_predict}, this leads to a covariance structure where off-diagonal elements of the noise covariance matrix are zero: $k_\nu(\Xi, \Xi')=0$ for $\Xi \neq \Xi'$. The diagonal elements, representing the variance at each input point, are modeled as $k_\nu(\Xi, \Xi) = \sigma_\nu^2(\Xi)$.

We then model the noise variance $\sigma_\nu(x)$ using the following formulation:
\begin{equation}
\label{formu:noise_variance}
    \sigma_\nu(\Xi) = z \cdot \exp\left( \beta^\top \rho(\Xi) \right) + \zeta,
\end{equation}
where $\beta \in \mathbb{R}^m$ represents the regression coefficients, $\zeta \geq 0$ is the minimum noise variance, the scale factor $z$ controls the overall magnitude of the noise variance, and the function $\rho(\Xi)$ is a polynomial feature map applied to the input $\Xi$. This noise model allows for dynamic adjustment of the noise variance across the search space, thereby enabling the Gaussian process to account for varying levels of uncertainty at different input points during the optimization process.

After defining the parametric form of $\sigma_\nu$, we estimate the noise $\nu$ via Gaussian process regression. First, we fit the GP to the objective function to predict the error $\hat{e}(\Xi)$. The residual noise is then computed as: $\nu(\Xi):=\left| e(\Xi) - \hat{e}(\Xi) \right|$, where $e(\Xi)$ denotes the observed error. Subsequently, we fit the noise model $\epsilon$ to these noise values $\nu(\Xi)$ as a regression problem, which enables us to estimate the noise variance $\sigma_\nu(\Xi)$ for adjusting our GP model.

The complete HBO process iteratively adjusts the PID parameters by minimizing the tracking error function:
\begin{equation}
    \boldsymbol{\Xi}^* = \arg\min_{\boldsymbol{\Xi}} e(\mathbf{\Xi}),
\end{equation}
where the tracking error $e$ is defined as:
\begin{equation}
    e=mean(e_{p})+\alpha \cdot mean(e_{\psi}),
\end{equation}
where $e_{p}$ represents the position error, $e_{\psi}$ is the angular error, and weight $\alpha$ determines the relative importance of each term in the error function. 

The overall HBO algorithm is summarized in Algorithm \ref{alg:hbo}, where the EI acquisition function is employed to guide the optimization process. The Expected Improvement quantifies the expected reduction in tracking error relative to the current best observation $e^*$, and is defined as:



\begin{equation}
\label{formu:EI}
\scalebox{0.92}{$
    EI(\Xi) = \left( e^* - \mu(\Xi) \right) \Phi\left( \frac{e^* - \mu(\Xi)}{\sigma(\Xi)} \right) + \sigma(\Xi) \phi\left( \frac{e^* - \mu(\Xi)}{\sigma(\Xi)} \right),
$}
\end{equation}
where $\mu(\Xi)$ and $\sigma(\Xi)$ denote the posterior mean and standard deviation of the heteroscedastic GP model. $\Phi(\cdot)$ and $\phi(\cdot)$ represent the standard normal cumulative distribution function and probability density function, respectively. The EI function enables the optimization process to find the global optimal solution more efficiently by embedding the input-dependent noise variance $\sigma_\nu(\Xi)$ into the a posteriori uncertainty $\sigma(\Xi)$, allowing the optimization process to dynamically adjust the weighting of the exploration and exploitation according to the noise level.

\subsection{Phased Optimization Strategy}
In the optimization of controller parameters, we adopted a phased optimization approach to effectively handle the system's behavior at different stages. This strategy gradually adjusts the controller parameters based on the error characteristics in the early and later stages, improving the system's overall performance and stability. To ensure robust performance under dynamic conditions, we further applied HBO to dynamically tune the PID gains, optimizing them based on time-varying tracking errors. This approach is particularly effective in the presence of heteroscedastic noise, where the variance of the noise varies with the system's state or input parameters. The adaptive nature of this strategy allows the PID controller to adjust to changing error dynamics and external disturbances, ensuring optimal tracking performance across varying system dynamics and noise levels.

\subsubsection{Phase 1: Short-Term Optimization}
In the first stage of optimization, we mainly focus on initial tuning of the controller parameters. At this time, the error fluctuations of the system are large, especially in the initial trajectory tracking process, the controller has not yet fully converged. Therefore, in this stage, we use Bayesian optimization to quickly determine the initial parameter settings of the controller. The goal of this stage is to quickly reduce the large error, stabilize the behavior of the system gradually, and find a suitable range of controller parameters to lay the foundation for subsequent fine-tuning.

\subsubsection{Phase 2: Long-Term Optimization}
When entering the second stage, the controller's preliminary parameter values obtained in the first stage are used as the starting points for the second stage optimization. These parameters are further fine-tuned through HBO to better adapt to the subtle changes in the system. Although the error of the system is relatively small at this stage, there are still some small fluctuations, especially in the tracking accuracy of attitude angles. Therefore, the optimization in the second phase focuses on reducing these small errors to ensure that the system can operate stably in complex environments.

Through this phased optimization strategy, we can avoid the computational waste of fine-tuning too early and ensure the optimal selection of controller parameters. 
The complete procedures of optimizing the controller hyperparameters are shown in Algorithm~\ref{alg:hbo}. 
At each phase, we update the GP model and the noise model, adjusting the model's noise predictions based on the newly observed data. This ensures that the optimization process adapts to the varying noise levels observed in different regions of the parameter space, leading to better performance and more robust optimization of the controller hyperparameters.

\begin{algorithm}[t]
\caption{HBO-based PID Control Algorithm}
\begin{algorithmic}[1]
\State \textbf{Input:}
Controller parameter search space $\mathcal{S} \subset \mathbb{R}^3$, number of HBO iterations $\bar{n}$, simulation steps $n$, initial sampling number $\check{n}$, reference trajectory $\mathcal{T}_{\text{ref}}$
\State \textbf{Output:}
Optimal parameters ${\Xi}^*$
\State Build GP model $\mathcal{GP}$ and noise model $\mathcal{\epsilon}$
\State Sample initial parameters $\{\Xi_i\}_{i=1}^{\check{n}} \in \mathcal{S}$
\State Get errors $\{e_i\}_{i=1}^{\check{n}}=f(\Xi_i, \mathcal{T}_{\text{ref}})$ for i in $\check{n}$
\State Build sample dataset $\mathcal{D} = \{ \mathbf{\Xi}, \mathbf{e} \} = \{(\Xi_i, e_i)\}_{i=1}^{\check{n}}$
\State Fit model $\mathcal{GP}$ on $\mathcal{D}$
\For{$j = 1$ \textbf{to} $\bar{n}$}
    \State Fit noise model $\epsilon$ using residuals $|\mathbf{e} -\mathcal{GP}(\mathbf{\Xi)}|$
    \State Update the noise variance of $\mathcal{GP}$ using $\sigma_{\nu} = \epsilon(\mathbf{\Xi})$
    \State Retrain $\mathcal{GP}$ on $\mathcal{D}$
    \State Get the next sample point $\Xi_{j}=EI(\mathcal{GP},e^*)$
    \For{$k = 1$ \textbf{to} $n$}
        \State $a_k^j = \text{PID}(\mathbf{\Xi}_j)$ \Comment{PID controller action}
        \State $e_k^j = \text{SendToActuators}(a_k^j)$ \Comment{Tracking error}
    \EndFor
    \State Get the average error $e_j=\frac{1}{n}\sum_{k=1}^{n} e_k^j(\Xi_j)$
    \State Update sample dataset $D = D \cup \{(\Xi_j, e_j)\}$
\EndFor
\end{algorithmic}
\label{alg:hbo}
\end{algorithm}

\renewcommand{\arraystretch}{1.0}
\begin{table}[t]
\caption{Physical parameters of quadrotors used in simulation.}
\small 
\label{table_example}
\begin{center}
\begin{tabular}{c c c c}
\hline
Variable & Symbol & Value & Unit \\
\hline
Mass & $m$ & 27 & $\text{g}$ \\
\hline
Gravity & $g$ & 9.81 & $\text{m}/\text{s}^2$ \\
\hline
Arm length & $L$ & 3.97 & $\text{cm}$ \\
\hline
Inertia along x-axis & $J_x$ & $1.40 \times 10^{-5}$ & $\text{kg} \, \text{m}^2$ \\
\hline
Inertia along y-axis & $J_y$ & $1.40 \times 10^{-5}$ & $\text{kg} \, \text{m}^2$ \\
\hline
Inertia along z-axis & $J_z$ & $2.17 \times 10^{-5}$ & $\text{kg} \, \text{m}^2$ \\
\hline
Thrust coefficient & $C_f$ & $2.88 \times 10^{-8}$ & $\text{kg} \, \text{m}/\text{rad}^2$ \\
\hline
Torque coefficient & $C_\tau$ & $7.24 \times 10^{-10}$ & $\text{kg} \, \text{m}^2/\text{rad}^2$ \\
\hline
Drag coefficient & $T_m$ & 0.02 & $\text{s}$ \\
\hline
Motor coefficient & $k_m$ & $2.81 \times 10^{-2}$ & rad \\
\hline
Motor bias & $b_m$ & 426.24 & $\text{rad}/\text{s}$ \\
\hline
Maximum PWM & $h_{max}$ & 65535 & $\text{Hz}$ \\
\hline
\end{tabular}
\end{center}
\label{tab:table1}
\end{table}

\section{Experiments and Results}
In this experiment, we compared the proposed HBO-PID with several baseline methods to evaluate their control performance in both simulation and real-world experiments. The considered baseline methods include traditional PID~\cite{lopez2023pid}, LQR~\cite{foehn2018onboard}, MPC~\cite{nan2022nonlinear}, TD3~\cite{han2023symmetric}, PID based on random search (RS-PID), Bayesian Optimization-based PID (BO-PID)~\cite{berkenkamp2016safe}. Inspired by previous work ~\cite{han2023symmetric}, we optimized the model parameters of all controllers in a simulation environment to ensure the rationality of the experimental design and the fairness of the comparison.







\subsection{Experimental Setup}
The goal of the experiments is to simulate the trajectory tracking of a UAV using a combination of position and attitude controllers. The simulation was implemented using Python 3.8 and PyBullet~\cite{coumans2016pybullet}, a widely used physics engine for robotics simulations.

\subsubsection{UAV Model and Parameters}
The UAV used in the simulation is a quadrotor model, which is a lightweight drone suitable for both research and practical applications. The physical parameters of the UAV, including its mass, moment of inertia, and maximum thrust, are summarized in Table~\ref{tab:table1}. These parameters are crucial for accurately modeling the dynamic behavior of the UAV during trajectory tracking.

\begin{figure*}[thpb]
    \centering
    \includegraphics[width=0.9\linewidth]{figure/Trajectory.png}
    \caption{Tracking results of different algorithms on three trajectories, ellipse on the left, four-leaf clover in the center, and spiral trajectory on the right. As shown by the red arrows, our method is consistently closer to the reference trajectory in all three trajectories.}
    \label{fig:tracking}
\end{figure*}

\subsubsection{Simulation Settings}
The simulation was executed with a time step of $\Delta t = 0.01$s, which provides a balance between computational efficiency and simulation accuracy. To ensure realistic behavior, the acceleration limits are set as follows:
\begin{itemize}
    \item Translational acceleration limit: $c_t = 5 \, \text{m/s}^2$.
    \item Rotational acceleration limit: $c_r = 20 \, \text{rad/s}^2$.
\end{itemize}
These constraints reflect typical operational limits for small UAVs under nominal conditions and prevent unrealistic acceleration values during the simulation.

\subsubsection{Hardware Environment}
The experiments were conducted on a personal computer (PC) equipped with an Intel i5-13500H CPU and 16 GB of RAM. No GPU was used for this simulation, ensuring that the results are applicable to systems with limited computational resources. The use of a CPU-only setup emphasizes the practicality of the proposed methods in real-world scenarios where GPU acceleration may not always be available.

\subsubsection{Error Metrics}
To evaluate the performance of the trajectory tracking, the average position error $e_p$ and angular error $e_\psi$ over the duration of the simulation are computed as follows:
\begin{equation}
    \begin{aligned}
    e_p & = \frac{1}{T}\int_0^T \|p(t)-p_d(t)\| \, dt, \\
    e_\psi & = \frac{1}{T}\int_0^T |\psi(t)-\psi_d(t)| \, dt,
    \end{aligned}
\end{equation}
where $p(t)$ and $\psi(t)$ denote the actual position and angle of the UAV, and $p_d(t)$ and $\psi_d(t)$ represent their desired values, respectively.

\begin{table*}[t]
    \centering
    \caption{The performance comparison of different control algorithms across three trajectories. (Note: \( e_p \) represents position error in meters (\si{m}), \( e_\psi \) represents heading error in degrees (°).)}
    \small 
     \resizebox{0.8\textwidth}{!}
     {
    \begin{tabular}{ccccccc}
        \toprule
        \textbf{} & \multicolumn{2}{c}{\textbf{Ellipse}} & \multicolumn{2}{c}{\textbf{Four-leaf clover}} & \multicolumn{2}{c}{\textbf{Spiral}} \\
        \midrule
        \textbf{Controller} & $e_p$(m) & $e_\psi$($^\circ$) & $e_p$(m) & $e_\psi$($^\circ$) & $e_p$(m) & $e_\psi$($^\circ$) \\
        \midrule
        PID & 0.233$\pm$0.223 & 1.499$\pm$0.731 & 0.302$\pm$0.072 & 6.723$\pm$3.264 & 0.588$\pm$0.186 & 2.465$\pm$2.718 \\
        \midrule
        LQR & 0.240$\pm$0.224 & 1.261$\pm$0.613 & 0.306$\pm$0.074 & 5.668$\pm$2.749 & 0.593$\pm$0.185 & 2.086$\pm$2.612 \\
        \midrule
        MPC & 0.234$\pm$0.225 & 1.471$\pm$0.716 & 0.300$\pm$0.072 & 6.588$\pm$3.189 & 0.583$\pm$0.187 & 2.412$\pm$2.673 \\
        \midrule
        TD3 & 0.182$\pm$0.231 & 0.547$\pm$0.471 & 0.232$\pm$0.065 & 2.762$\pm$1.475 & 0.491$\pm$0.198 & 1.046$\pm$2.726 \\
        \midrule
        RS-PID  & 0.226$\pm$0.253 & 1.406$\pm$0.703 & 0.233$\pm$0.074 & 6.249$\pm$3.033 & 0.387$\pm$0.270 & 2.303$\pm$2.666 \\
        \midrule
        BO-PID  & 0.185$\pm$0.245 & 0.778$\pm$0.428 & 0.204$\pm$0.067 & 3.476$\pm$1.684 & 0.684$\pm$0.696 & 1.852$\pm$3.107 \\
        \midrule
        HBO-PID (Ours)  & \textbf{0.137$\pm$0.233} & \textbf{0.323\bm{$\pm$}0.397} & \textbf{0.162\bm{$\pm$}0.052} & \textbf{1.328\bm{$\pm$}1.260} & \textbf{0.305\bm{$\pm$}0.226} & \textbf{0.649\bm{$\pm$}2.612} \\
        \bottomrule
    \end{tabular}
    \label{tab:table2}
    }
\end{table*}

\begin{figure}[t]
    \centering
    \includegraphics[width=1.0\linewidth]{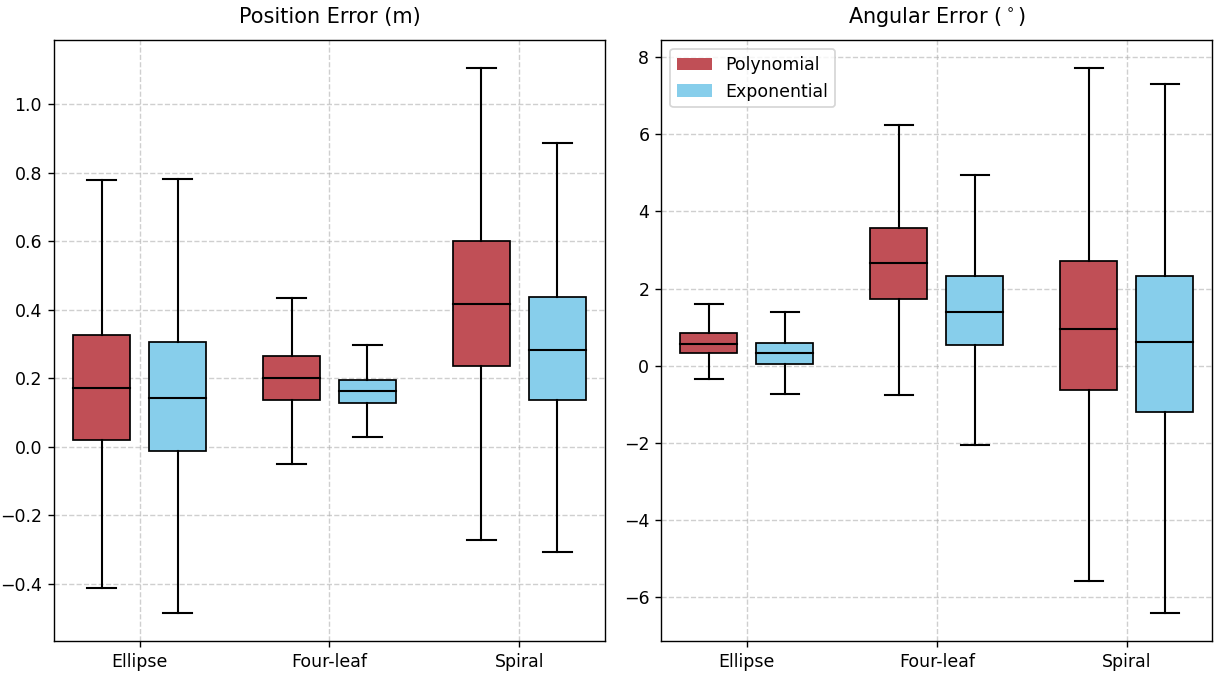}
    \caption{Tracking error of using different noise models.}
    \label{fig:Compare Noise}
\end{figure}

\begin{figure}[t]
    \centering
    \includegraphics[width=1.0\linewidth]{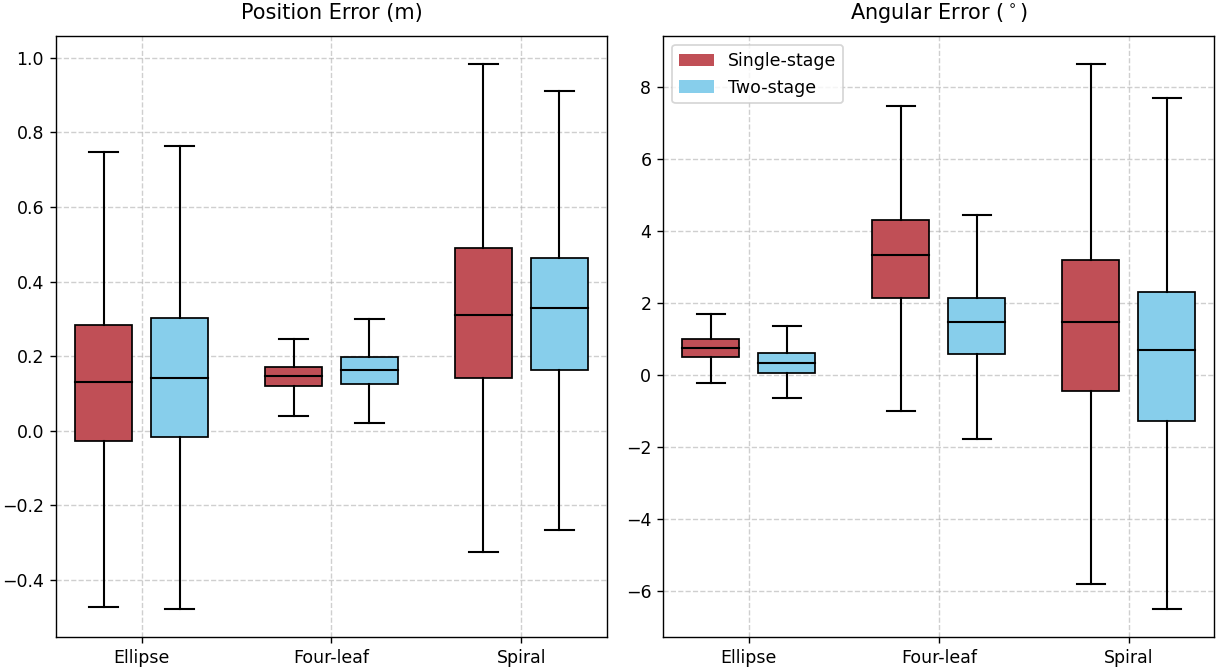}
    \caption{Tracking error of using different optimization strategies.}
    \label{fig:Compare Two-Stage}
\end{figure}

\begin{figure*}[t]
    \centering
    \includegraphics[width=0.9\linewidth]{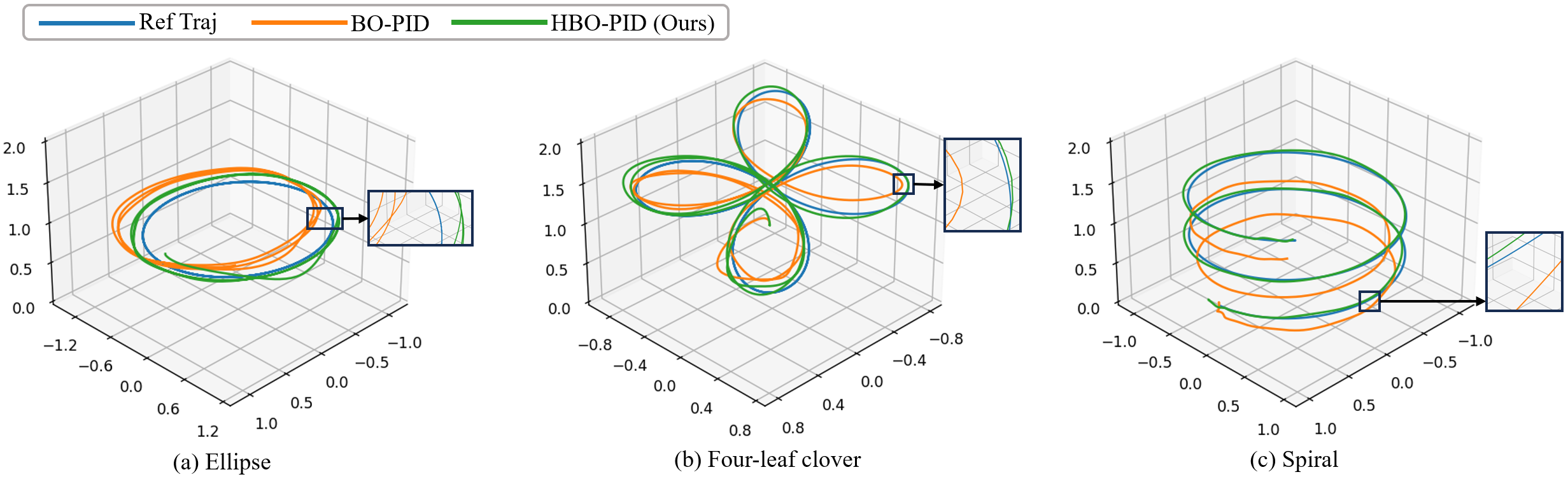}
    \caption{Comparison of trajectories generated by using different controllers in real-world experiments.}
    \label{fig:real_trajectory}
\end{figure*}

\subsection{Trajectory Tracking Performance}
To evaluate the performance of HBO-PID, three distinct trajectories were designed for the UAV trajectory tracking. Each trajectory lasted for $T$ = 50 s and was chosen to test the UAV’s ability to track different types of motion, including smooth and more dynamic variations in both position and attitude. The trajectories were designed to challenge the PID controller's performance, ensuring that it could handle varying levels of complexity in the tracking task.

For each of these trajectories, the UAV was set to track the desired position and attitude. The position and attitude information at each time step were provided to the controller, which was responsible for adjusting the UAV’s control inputs. Fig~\ref{fig:tracking}. compares the tracking results of using various control algorithms on these three different trajectories, and it can be seen that the proposed HBO-PID performs the best.

The results of different controllers are demonstrated in Table~\ref{tab:table2}. It can be seen that the proposed method (HBO-PID) outperforms all the baseline methods in terms of position error and angular error in all the tested trajectories. Specifically, HBO-PID achieves a position error of 0.137 m on Ellipse trajectory, which is 24.7\%, 25.9\%,  39.4\%, 41.2\%, 41.2\%, and 42.9\% lower than the TD3 (0.182 m), BO-PID (0.185 m), RS-PID (0.226 m), PID (0.233 m), MPC (0.234 m), and LQR (0.240 m), respectively. The improvement of HBO-PID over other methods is more significant, ranging from 40.9\% to 78.4\% on Ellipse. It is the same trend on the other two trajectories. We can also find that the introduction of the nonlinear dynamic compensation mechanism can well improve the performance of the classical PID controller, which is justified by the fact that both RS-PID and BO-PID perform better on all the trajectories.

\subsection{Ablation Experiment}
In order to validate the effectiveness of the noise modeling and two-stage optimization strategy, we conducted an ablation study on the three trajectories. 

\subsubsection{Exponential vs. Polynomial Noise Model}
We compare the results of using different noise model for HBO. Fig.~\ref{fig:Compare Noise} shows that the exponential noise model outperforms polynomial noise model across all trajectories. For instance, on spiral trajectory, the exponential model reduces the position error by 26.5\% and angular error by 39.3\%. This implies that the exponential noise model can better capture the variations in the noise variance across the search space, whereas the polynomial model struggles to generalize under highly nonlinear noise patterns.

\subsubsection{Two-Stage vs. Single-Stage Optimization}
As shown in Fig.~\ref{fig:Compare Two-Stage}, the two-stage optimization strategy significantly reduces angular errors compared to the single-stage approach. On four-leaf clover trajectory, the angular error decreases from $3.342^\circ$ (single-stage) to $1.328^\circ$ (two-stage), reaching an error reduction of 60.3\%. Similarly, the spiral trajectory exhibits an improvement of 52.6\% in angular accuracy. While the position error slightly increases in some cases, the total weighted error $e_{\text{total}} = e_p + \alpha \cdot e_{\psi}$ remains lower due to the emphasis on angular precision in the second stage. This phased strategy avoids premature fine-tuning and enables robust adaptation to dynamic error characteristics.

\subsection{Real-world experiment}

\begin{table}[!t]
    \centering
    \caption{The performance comparison of different algorithms across three trajectories.}
    \small 
     \resizebox{1.0\linewidth}{!}
     {
    \begin{tabular}{ccccccc}
        \toprule
        \textbf{Trajectory} & \multicolumn{2}{c}{\textbf{Ellipse}} & \multicolumn{2}{c}{\textbf{Four-leaf clover}} & \multicolumn{2}{c}{\textbf{Spiral}} \\
        \midrule
        \textbf{Error} & $e_p$(m) & $e_\psi$($^\circ$) & $e_p$(m) & $e_\psi$($^\circ$) & $e_p$(m) & $e_\psi$($^\circ$) \\
        \midrule
        BO-PID & 0.368 & 0.662 & 0.442 & \textbf{0.503} & 0.541 & 0.591 \\
        \midrule
        HBO-PID (Ours) & \textbf{0.141} & \textbf{0.616} & \textbf{0.382} & 0.558 & \textbf{0.318} & \textbf{0.383} \\
        \bottomrule
    \end{tabular}
    \label{error:table3}
    }
\end{table}

\begin{figure*}[t]
    \centering
    \includegraphics[width=1.0\linewidth]{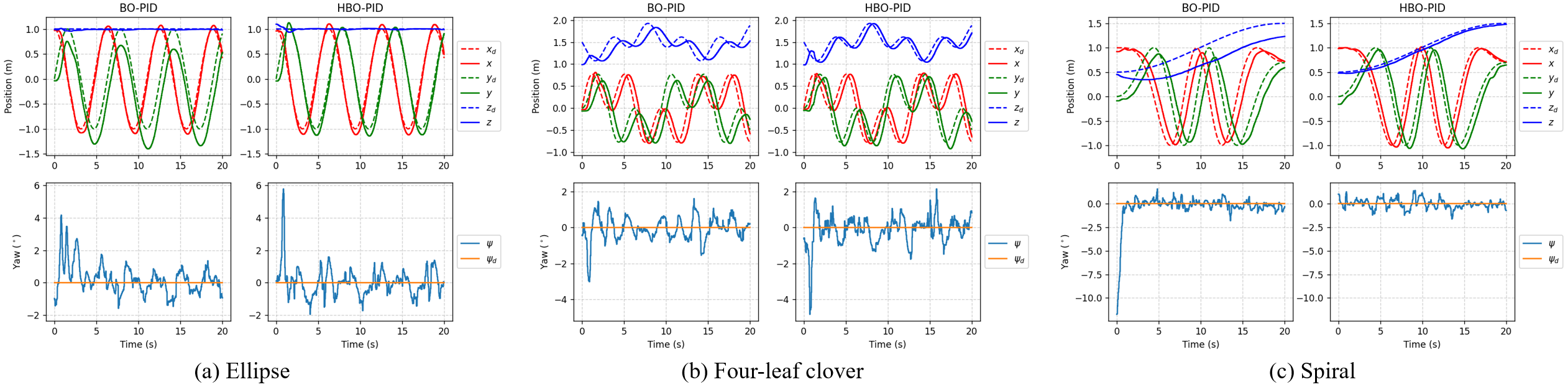}
    \caption{Comparison of tracking performance in real-world experiments. In the upper half of each subfigure, the red curve shows the UAV position trajectory in the x-direction, the green curve in the y-direction, the blue curve in the z-direction and the lower half shows the actual variation of the yaw angle $\psi$, where the desired yaw angle $\psi_d$ is set to 0 to keep the heading stable during the flight.}
    \label{fig:real_trajectory_table}
\end{figure*}

To validate the proposed method, we carried out real-world experiments using a quadrotor UAV equipped with a nonlinear controller, operating under the NOKOV motion capture system.
The quadrotor UAV used in the experiments had a mass of 800 grams(g), a maximum thrust of 2.99 Newtons(N), and a minimum thrust of 0.22 Newtons(N). The controller parameters were optimized using the approach described in this work. The experimental platform was built on the open-source hardware OminiNxt, and the Px4 controller was employed to track the commanded collective thrust and body rates. The experiments were conducted within a motion capture system with a measurement volume of $6 \times 4 \times 3$ meters. The system operated at a frequency of 100 Hz, providing accurate position and attitude measurements for the UAV during the experiments.

We developed a UAV trajectory tracking system based on the HBO-PID controller. Three trajectory patterns—ellipse, four-leaf clover, and spiral—were replicated in real-world flight scenarios, maintaining geometric isomorphism with their simulated counterparts (vertical scale compression was applied due to spatial constraints). We compare the results of HBO-PID with those of its counterpart BO-PID. As can be seen in Table~\ref{error:table3}, the proposed HBO-PID achieves remarkable improvements in tracking accuracy and robustness. Compared to BO-PID, HBO-PID reduces the mean position error by 61.2\%, and the mean attitude angle by 34.7\%. Fig.~\ref{fig:real_trajectory} visually compares the tracking performance across the three trajectories, which suggests that the trajectory generated by using the HBO-PID is much closer to the ground truth than BO-PID. Fig.~\ref{fig:real_trajectory_table} provides more details about the position error and angular error of trajectory tracking, from which we can see that HBO-PID significantly behaves better than BO-PID.

\section{Conclusions}
In this paper, we propose a novel HBO-PID for accurate and robust trajectory tracking of UAVs. By leveraging Gaussian processes to explicitly model the noise variance correlated with the PID controller inputs, HBO-PID significantly improves the accuracy and robustness of trajectory tracking in dynamic and complex environments. Besides, it utilizes a two-stage optimization strategy, which effectively accelerates the convergence of parameter tuning, enabling the controller to achieve optimal performance more efficiently. Experimental results in both simulation and real-world experiments demonstrate that HBO-PID outperforms SOTA methods in both the position error and angular error. In the future, we will work on further improving the computational efficiency of HBO, extending it to more practical tasks. 

\addtolength{\textheight}{-12cm}   







\bibliographystyle{ieeetr}
\bibliography{IEEEfull}

\end{document}